\documentclass[letterpaper, 10 pt, conference]{ieeeconf}
\IEEEoverridecommandlockouts

\usepackage{graphics}
\usepackage{graphicx}
\usepackage{epsfig}
\usepackage{multirow}
\usepackage{amssymb}
\usepackage{hyperref}
\usepackage{subcaption}
\usepackage{amsmath}
\usepackage{booktabs}
\usepackage{longtable}
\usepackage{stackengine}
\usepackage{xfrac}
\usepackage{tabularx}
\usepackage{multirow}
\usepackage{cite}
\usepackage{float} 
\usepackage{cancel}
\usepackage[normalem]{ulem} 
\usepackage{bm}
\usepackage{balance}
\usepackage[vlined,boxruled]{algorithm2e}  

\makeatletter
\def\old@comma{,}
\catcode`\,=13
\def,{%
  \ifmmode%
    \old@comma\discretionary{}{}{}%
  \else%
    \old@comma%
  \fi%
}
\makeatother

\usepackage[dvipsnames]{xcolor,colortbl}
\usepackage{bm} 
\newtheorem{remark}{Remark}

\newcommand{\R}{\mathbb{R}}

\newtheorem{prob}{Problem}

\begin{document}

\graphicspath{{Figures/}}
\allowdisplaybreaks

\title{\LARGE \bf Application-Oriented Co-Design of Motors and Motions for a 6DOF Robot Manipulator}

\author{Adrian Stein, Yebin Wang, Yusuke Sakamoto, Bingnan Wang, and Huazhen Fang
\thanks{A. Stein is with the Department of Mechanical and Aerospace Engineering, University at Buffalo, NY 14260, USA. This work was done while he was a research intern at Mitsubishi Electric Research Laboratories. {\tt\small(email: astein3@buffalo.edu)}.}
\thanks{Y. Wang, Y. Sakamoto, and B. Wang are with Mitsubishi Electric Research Laboratories, Cambridge, MA 02139, USA. {\tt\small (email: \{yebinwang,sakamoto,bwang\}@merl.com)}.}
\thanks{H. Fang is with the Department of Mechanical Engineering, University of Kansas, Lawrence, KS 66045, USA. {\tt\small(email: fang@ku.edu)}.}
}

\maketitle

\begin{abstract}
This work investigates an application-driven co-design problem where the motion and motors of a six degrees of freedom robotic manipulator are optimized simultaneously, and the application is characterized by a set of tasks. Unlike the state-of-the-art which selects motors from a product catalogue and performs co-design for a single task, this work designs the motor geometry as well as motion for a specific application. Contributions are made towards solving the proposed co-design problem in a computationally-efficient manner. First, a two-step process is proposed, where multiple motor designs are identified by optimizing motions and motors for multiple tasks one by one, and then are reconciled to determine the final motor design. Second, magnetic equivalent circuit modeling is exploited to establish the analytic mapping from motor design parameters to dynamic models and objective functions to facilitate the subsequent differentiable simulation. Third, a direct-collocation-based differentiable simulator of motor and robotic arm dynamics is developed to balance the computational complexity and numerical stability. Simulation verifies that higher performance for a specific application can be achieved with the multi-task method, compared to several benchmark co-design methods. 
\end{abstract}


\IEEEpeerreviewmaketitle

\section{Introduction}\label{sec:introduction}
Off-the-shelf industrial robotic manipulators usually come with specifications to fulfill the requirements of a broad range of customers. Their designs are often optimized for general-purpose applications~\cite{Toussaint.2021}, which implies sub-optimality for a specific application. Application-oriented robot design, which optimizes robot for a specific application, offers a great potential to deliver more cost-effective solutions, e.g., higher productivity, less energy consumption, and lower initial cost.
Robot design is multidisciplinary in nature, involving battery~\cite{Findlay.2015}, structural mechanics~\cite{Zhou.2015}, kinematics~\cite{Mayorga.2005}, dynamics~\cite{Zhou.2011}, thermodynamics~\cite{PadillaGarcia.2018}, and control~\cite{Kwok.1994}. The design objectives are also multidisciplinary, e.g.,  weight~\cite{Zhou.2011,Zhou.2015,Findlay.2015,Li.2021,PadillaGarcia.2018}, energy consumption~\cite{PadillaGarcia.2018,BravoPalacios.2020,Stevo.2014}, task completion time~\cite{BravoPalacios.2020,Stevo.2014} or workspace maximization~\cite{Li.2021}. Much of the established robot design methods have been discipline-specific. Some typical examples include, kinematic design optimization with a posture determination~\cite{Mayorga.2005}, kinematic design to optimize the position and orientation for a specific operation~\cite{Raza.2018}, multi-objective design for workspace, path planning or lightweight optimization~\cite{Castejon.2010,Bugday.2019}, and PID gain tuning for system performance~\cite{Kwok.1994}. Co-design, which reconciles the coupling and conflict of subsystems at an early stage~\cite{PadillaGarcia.2018}, is of interest for its potential to overcome sub-optimality that results from a discipline-specific design process. Its use can be found in applications such as  manipulators~\cite{Toussaint.2021,Bastos.2015,Pettersson.2009,Zhou.2011,Zhou.2015}, legged robots~\cite{Dinev.2022,Ha.2018}, aerial/ground manipulators~\cite{Findlay.2015}, service robotic arms~\cite{Li.2021} and industrial robots~\cite{PadillaGarcia.2018}, just to name a few. Particularly, the work~\cite{Pettersson.2009} co-designs the drivetrain and joint trajectories of an industrial robot, where the drivetrain is parameterized in terms of the motor shaft length and gearbox ratio. The works~\cite{Bastos.2015,BravoPalacios.2022} simultaneously optimize trajectories and controllers. The works~\cite{Zhou.2011,Findlay.2015} take the gearbox or motor selection into account. One of the main challenges in co-design is the heavy computational burden~\cite{BravoPalacios.2022}. Consequentially, most co-design work restricts the number of axis or motor design freedom~\cite{Pettersson.2009,Ravichandran.2006}.

This paper proposes a systematic modeling-to-computation pipeline to co-design the motions (joint trajectories) and actuators of a six degrees of freedom (6DOF) robotic manipulator to improve productivity or energy efficiency for a considered application. Each actuator is assumed to be a surface permanent magnet synchronous motor (SPMSM) and the application is characterized by a set of tasks. The co-design problem differs from the state-of-the-art in that the motor design involves parameters representing the geometric shape other than selecting motors from a product catalogue. The proposed co-design process tackles the computational challenge through employing: 1) magnetic equivalent circuit (MEC) modeling to establish the analytic mapping from motor design to dynamic model parameters; 
2) a differentiable simulator of motor and arm dynamics based on CasADi~\cite{Andersson.InPress2018} where the analytical formula of the gradient and Hessian can be derived via auto-differentiation; 3) direct collocation-based integration of motor and arm dynamics to balance accuracy and efficiency; and 4) a two-step process to trade-off between efficiency and optimality, where the motion and motor design is solved for each task, and then the candidate motor designs corresponding to all tasks are projected to the feasible region to reach the final motor design.


The notation throughout this paper follows~\cite{Lynch.2017}. Section~\ref{sec:system_modeling_and_problem_formulation} describes the system modeling and problem formulation. Section~\ref{sec:main_results} presents the main results. Section~\ref{sec:simulation} shows the simulation, and Section~\ref{sec:conclusions} concludes the work.
\section{System Modeling and Problem Formulation}
\label{sec:system_modeling_and_problem_formulation}
This section describes the modelling of an SPMSM and a 6DOF manipulator, and formulates the co-design problem. 
\subsection{SPMSM Design Parameterization and Modeling}
\label{subsec:design_of_the_SPMSM}
\begin{figure}[!t]
\centering
	\includegraphics[width=0.4\textwidth]{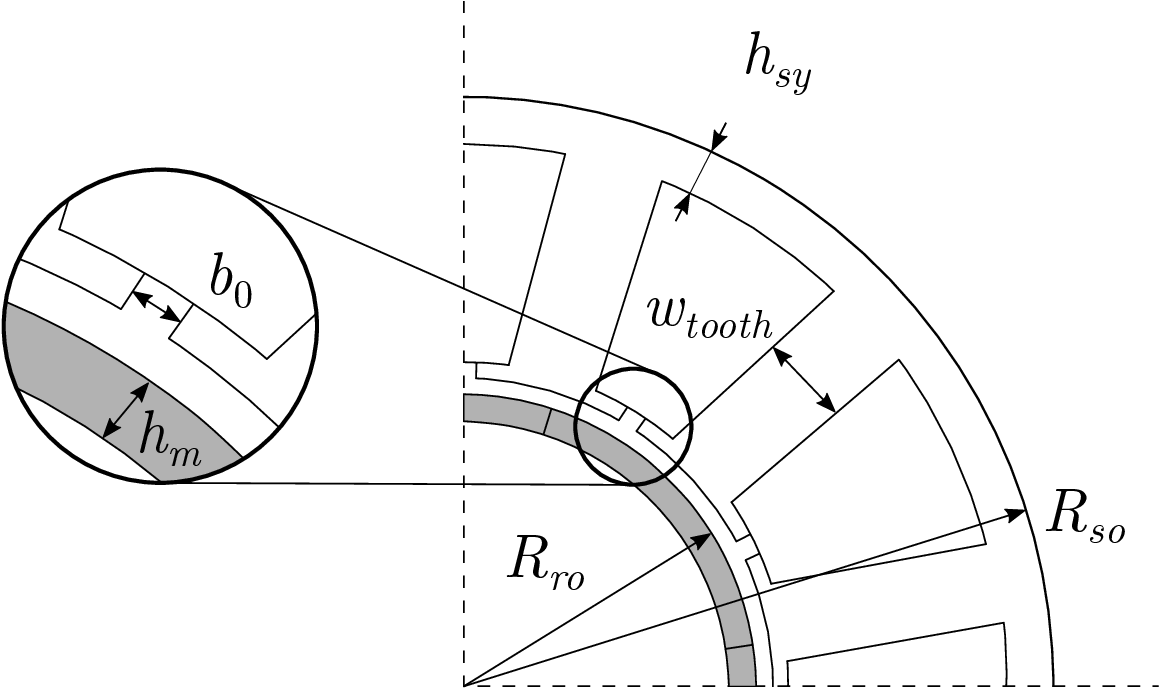}
	\caption{The cross-section of the magnetic design of the SPMSM. (Note: The axial stack length $L$ is not illustrated.)}
 \label{fig:0_Motor}
\end{figure}

Fig. \ref{fig:0_Motor} illustrates motor design variables to be determined. The physical meanings of these variables can be found in Table \ref{table:FEM_vs_analytical_results}. The following motor parameters have fixed values:
\begin{itemize}
    \item Number of pole pairs $P = 4$
    \item Number of slots $Q = 12$
    \item Height of tooth tip $h_{tip}=2$ mm
    \item Width of air gap $\delta=0.5$ mm
    \item Number of winding turns per tooth $n_s=50$
    \item Number of coils connected in parallel $C_p = 1$
    \item Remanent flux density of the magnet $B_r = 1.38$ T
    \item Relative recoil permeability of the magnet $\mu_r = 1.05$
    \item Filling factor $f_f = 0.55$
    \item Maximum limitation for flux density $B_{max} = 1.5$ T
    \item Magnet width in electric angle $\alpha_m = \pi$
\end{itemize}
With zero skewness assumption, a slot/pole ratio of $12/8$ yields a winding factor of $k_{w1}=0.866$. Given a motor design, the MEC modeling, detailed in the Appendix, follows to analytically calculate the dynamic model parameters as listed in Table~\ref{table:FEM_vs_analytical_results}. Table \ref{table:FEM_vs_analytical_results} shows that the dynamic model parameters from the MEC modeling are comparable to those obtained by applying a finite element method. 
\begin{table}[thpb]
\begin{center}
\caption{Motor design and comparison results.}\label{table:FEM_vs_analytical_results}
\begin{tabular}{cccc}
\multicolumn{3}{c}{\textbf{Motor design variables $\Xi$}}\\
\hline
Parameter & Value & Unit \\
\hline
Axial length of core $L$ & $20$ & mm\\
Outer radius of rotor $R_{ro}$ & $18$ & mm\\
Outer radius of stator $R_{so}$ & $30$ & mm\\
Height of magnet $h_m$ & $3$ & mm\\
Stator yoke $h_{sy}$ & $5$ & mm\\
Width of tooth $w_{tooth}$ & $7$ & mm\\
Slot opening $b_0$ & $2$ & mm\\
Number of winding turns per tooth $n_s$ & $30$ & -\\
\hline
\hline\\
\multicolumn{4}{c}{\textbf{Dynamic model parameter values from MEC and FEM}}\\
\hline
Parameters & FEM & MEC & Unit\\
\hline
Magnetic flux $\Phi_m$ & $0.0303$ & $0.0288$ & Wb\\
Inductance $L_{d}$ & $0.714$ & $0.698$ & mH \\
Inductance $L_{q}$ & $0.708$ & $0.698$ & mH\\
Resistance $R$ & - & $0.621$ & Ohm\\
\hline
\end{tabular}
\end{center}
\end{table}
The dynamic model of the SPMSM is given below~\cite{Lemmens.2015}: 
\begin{subequations}
\label{eq:electric_EOMs}
\begin{align}
    \frac{di_{d}}{dt} &= -\frac{R}{L_d}i_{d} + P\omega i_{q} + \frac{u_{d}}{L_{d}}\label{eq:SPMSM_dynamic_model_id}\\
    \frac{di_{q}}{dt} &= -\frac{R}{L_{q}}i_{q} - P\omega \left(i_{d} + \frac{\Phi_{m}}{L_{q}}\right) + \frac{u_{q}}{L_{q}},\label{eq:SPMSM_dynamic_model_iq}
\end{align}
\end{subequations}
where $i_d$ and $i_q$ are the currents in d- and q-axis, respectively, and $\omega$ is the rotor velocity. For an SPMSM, its d-axis and q-axis inductance are equal, i.e., $L_d = L_q$. The motor produces a torque $\tau = \frac{3}{2}P (L_di_{d} + \Phi_{m}) i_{q}$.
The motor design variables are subject to the following design constraints: 
\begin{subequations} \label{eq:motor_design_constraints}
    \begin{align}
        \left[20,\:10,\:10,\:1,\:5,\:5,\:1\right]& \nonumber\\
        \leq \left[L,\:R_{ro},\:R_{so},\:h_{m},\:h_{sy},\:w_{tooth},\:b_{0}\right] &\leq \nonumber\\
        \left[100,\:100,\:100,\:5,\:10,\:20,\:10\right] &\label{eq:geometric_constraints}\\
        \arcsin\left(\frac{w_{tooth}}{2(R_{ro}+\delta)}\right) + \arcsin\left(\frac{b_0}{2(R_{ro}+\delta)}\right) &\leq \frac{\pi}{Q} \label{eq:con_tooth_width}\\
        h_{ss}&>0 \label{eq:con_slot_height}\\
        m_{\mathrm{stator}} + m_{\mathrm{rotor}} &\leq 3 \label{eq:con_weight} \\
        D_{wire} &\geq 0.6, \label{eq:con_min_wire_dia}
    \end{align}
\end{subequations}
where the constraints in~\eqref{eq:con_tooth_width}-\eqref{eq:con_min_wire_dia} enforce the maximum tooth width, minimal slot height, maximum weight and minimal wire diameter of each motor. The motor is additionally subject to the following operational constraints:
\begin{subequations}\label{eq:motor_operational_constraints}
    \begin{align}
        -3 \text{A} &\leq i_{d} \leq 3 \text{A}, \quad
        -3 \text{A} \leq i_{q} \leq 3 \text{A} \label{eq:con_idq}\\
        -100 \text{V} &\leq u_{d} \leq 100 \text{V},\quad
        -100 \text{V} \leq u_{q} \leq 100 \text{V}\label{eq:con_udq} \\
        \frac{k_p\Phi_m}{w_{tooth}L} &\leq 1.5\text{ T}, \quad
        \frac{1}{\sqrt{3}}\frac{k_p\Phi_m}{h_{sy}L} \leq 1.5 \text{ T},\label{eq:con_magn_flux_yoke}
    \end{align}
\end{subequations}
where \eqref{eq:con_idq}-\eqref{eq:con_udq} restrict the currents and voltages in d- and q-axis, and~\eqref{eq:con_magn_flux_yoke} constrains the magnetic fluxes in the tooth and the stator yoke. The torque ripples are not modelled.
\subsection{$6$DOF Open-Chain Robot Manipulator}
\label{subsec:6DoF_robotic_manipulator}
According to \cite{Lynch.2017}, the general dynamic model for the robot manipulator can be written in the form of
\begin{equation}\label{eq:robot_dynamic_model}
    M(\bm{\theta}) \ddot{\bm \theta} + C(\bm \theta,\dot{\bm \theta})\dot{\bm \theta} + G(\bm \theta) = \bm \tau,
\end{equation}
where $\bm\theta = \begin{bmatrix}\theta_{L_1} & \dots & \theta_{L_6}\end{bmatrix}^\top \in \R^6, \dot{\bm\theta},$ and $\ddot{\bm\theta}$  are the angles, velocities, accelerations of all links, respectively, and $\bm\tau$ are torques applied on all links; $M,C$ and $G$ are the link inertia matrix, Coriolis forces, and gravitational force. The units of $\theta,\dot\theta,\ddot\theta$ are rad, rad/s, rad/s$^2$, respectively. To solve the forward dynamics, this work makes use of the articulated-body algorithm (ABA)~\cite{Featherstone.2008,Featherstone.2010,Featherstone.2010b}, as shown in Section~\ref{sec:ABA_algorithm_with_gearbox_and_rotor_inertia}. No tip force is assumed. The constraints on the angles of velocities are given by:
\begin{subequations}\label{eq:robot_constraints}
    \begin{align}
        -2\pi \leq &\theta_{L_1},\theta_{L_4}, \theta_{L_6} \leq 2\pi \label{eq:con_theta_L_1}\\
        -0.6\pi \leq &\theta_{L_2}, \theta_{L_3}, \theta_{L_5} \leq 0.6\pi \\
        -100\pi \leq &\dot{\theta}_{R_k} \leq 100\pi,\label{eq:con_theta_dots}
    \end{align}
\end{subequations}
where subscripts $L_k$ and $R_k$ stand for the $k$th link and rotor, respectively. Given the gearbox ratio $Z$ being the ratio of the rotor speed to the associated link speed, we have $\omega = \dot{\theta}Z$. 
\subsection{Problem Formulation}
\label{subsec:problem_formulation_and_cost_function}
To perform application-oriented co-design, we first characterize an application as a set of $n$ tasks: $\mathcal T \triangleq \{\mathcal T_1, \mathcal T_2,...,\mathcal T_n\}$. Take pick and place in warehouses as an example. A task $\mathcal T_i$ is uniquely represented by a tuple $(\bm x_0, \bm x_f, M_p)$, where $\bm x_0 = (\bm \theta_0,\dot{\bm \theta}_0),\bm x_f =(\bm\theta_f,\dot{\bm\theta}_f)$,  and $M_p$ are the initial state, the final state, and the inertia matrix of the payload, respectively. We thus have the co-design problem as follows.
\begin{prob}\label{prob:co-design}
    Given a task set $\mathcal T$, the motor design parameterization $\Xi$, the motor model and constraints \eqref{eq:electric_EOMs} - \eqref{eq:motor_operational_constraints}, and the manipulator model and constraints \eqref{eq:robot_dynamic_model}-\eqref{eq:robot_constraints}, determine the optimal control $u^*_i$ of task $\mathcal T_i$ for $1 \le i \le n$ and optimal motor design $\Xi^*$ for all tasks such that a certain cost function $J(u_1,\cdots,u_n,\Xi)$ is minimized.
\end{prob}

The control inputs $u$ of the motor and manipulator dynamics aggregates the voltages in d- and q-axis of all six axes; and $\Xi$ aggregates the motor design parameters of all six axes. For a time-optimal co-design problem, the cost function is \mbox{$J=\sum^n_{i=1} t_{f,i}$}, where $t_{f,i}$ is the final time of task $\mathcal T_i$. For energy-optimal co-design, the cost function is \mbox{$J= \sum^n_{i=1} \int^{T_{fi}}_0 u_i^T(t)u_i(t) \mathrm{d} t$} where $T_{f,i}$ is the final time for task $\mathcal T_i$. Problem~\ref{prob:co-design}, designed to determine $u$ as functions over a certain time interval,
is infinite-dimensional. Thanks to the high dimension of manipulator dynamics, optimal control theory, e.g., minimum principle or dynamic programming, is unlikely to be applicable. We adopt the well-established direct transcript \cite{Bet98,WanZhaBor15,ZhaWanZho19} to reduce Problem~\ref{prob:co-design} into a finite-dimensional numerical optimization problem.
\begin{remark}
If $\mathcal T$ only contains one task, i.e., $n=1$, Problem~\ref{prob:co-design} is similar to that considered in \cite{Pettersson.2009} except that Problem \ref{prob:co-design} allows more motor design freedom.
\end{remark}
\section{Main Results}\label{sec:main_results}
This section presents the proposed approach to Problem~\ref{prob:co-design}: two-step co-design process, a differentiable simulator, and ABA with gearbox and motor inertia. The analysis of computational complexity and convergence is briefly discussed.

\subsection{Two-Step Co-Design Process}\label{subsec:co_design_process}
When $n>1$, solving Problem \ref{prob:co-design} at one time
incurs much higher computation burden than solving for $(\hat u^*_i,\hat \Xi^*_i)$ individually $n$ times. When $n$ becomes large, solving Problem \ref{prob:co-design} in one shot gets computationally intractable. Therefore we propose a two-step process: 1) solve for $(\hat u^*_i,\hat \Xi^*_i)$ corresponding to task $\mathcal T_i, 1 \le i \le n$; 2) reconcile $(\hat \Xi^*_1,\dots,\hat \Xi^*_n)$ to obtain $\Xi^*$ and solve for $u^*_i$ based on $\Xi^*$.
\subsubsection{Co-Design for Task $\mathcal T_i$}
\begin{figure}[thpb]
\centering
	\includegraphics[width=0.4\textwidth]{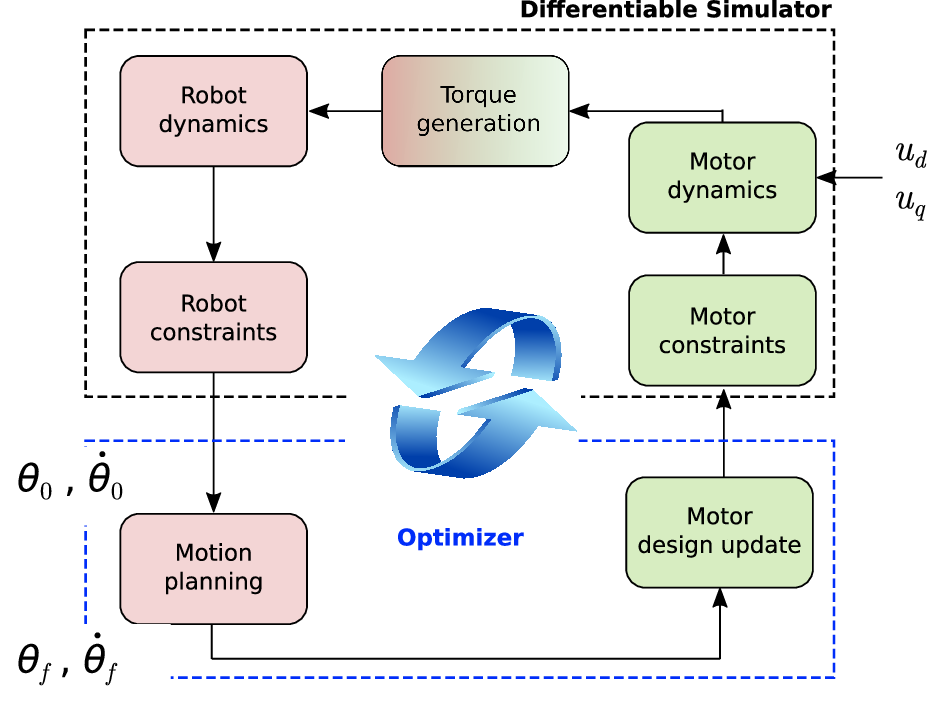}
	\caption{Co-design procedure for single task: the robot manipulator (in red) and the motor (in green).}
 \label{fig:01_Concept}
\end{figure}

The procedure is illustrated by Fig.~\ref{fig:01_Concept}, where the red and green boxes mark the manipulator and motor, respectively. It consists of two parts: 1) given $u$ and motor design $\Xi$, a differentiable simulator simulates the motor and manipulator dynamics to get state trajectory $\bm x(t), i_d(t), i_q(t)$ and calculates the gradient of the cost function w.r.t. $(u,\Xi)$; 2) the optimizer assesses the state trajectory against boundary conditions, dynamic and operational constraints, and if no violation occurs, updates control and motor design based on the gradient. The initial guess of $\Xi$ and $u$ is the mean of the upper and lower bound of the motor design variables as shown in~\eqref{eq:geometric_constraints} and~\eqref{eq:con_idq}-\eqref{eq:con_udq} respectively.
\subsubsection{Reconciliation of Motor Designs}
Applying the procedure as shown in Fig.~\ref{fig:01_Concept} to all tasks yields $n$ pairs of motion and motor: $(\hat u^*_i,\hat \Xi^*_i), 1 \le i \le n$. In a real world, only one motor design is adopted. Hence we need to come up with the motor design $\Xi^*$ based on $(\hat \Xi^*_1,\dots,\hat \Xi^*_n)$ systematically. 

\begin{figure}[thpb]
\centering
	\includegraphics[width=0.48\textwidth]{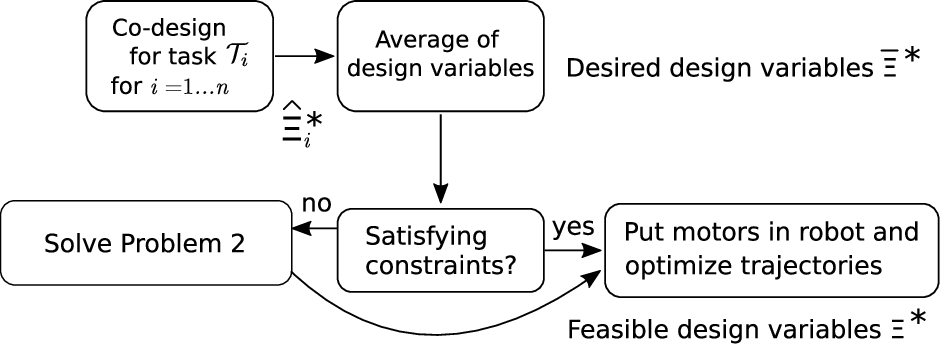}
	\caption{Reconciliation to obtain a unique motor design.}
 \label{fig:02_Concept_Co_Design}
\end{figure}

The reconciliation procedure is illustrated by Fig.~\ref{fig:02_Concept_Co_Design}.
Assume that we have $(\hat u^*_{i},\hat \Xi^*_{i})$ for $\mathcal T_i, 1 \le i \le n$. Then, the motor designs for all cases are weighted equally to get the average value $\bar{\Xi}^*$. 
Note that even if the motor design $\hat \Xi^*_{i}$ satisfies the constraints~\eqref{eq:motor_design_constraints} and~\eqref{eq:con_magn_flux_yoke}, $\bar{\Xi}^*$ might not, and the following 
constrained optimization problem needs to be solved to obtain a feasible motor design:

\begin{prob}\label{prob:reconcilation}
Find the feasible motor design $\Xi^*$ which is closest to the mean value $\bar{\Xi}^*$ by solving
\begin{equation*}
    \Xi^* = \arg\min_{\Xi \in \Omega} \sum_{k=1}^{n_\Xi} \frac{\left(\bar{\Xi}^*_{k} - \Xi_{k}\right)^2}{\left(\bar{\Xi}^*_k\right)^2},
\end{equation*}
where $n_\Xi$ is the dimension of $\Xi$; $\bar{\Xi}^*_k, \Xi_k$ are the $k$th element of the vectors $\bar{\Xi}^*$ and $\Xi$, respectively, and $\Omega$ is the feasible domain of $\Xi$ induced by constraints~\eqref{eq:motor_design_constraints} and~\eqref{eq:con_magn_flux_yoke}.
\end{prob}

If all elements in $\Xi^*$ are independent from each other and $\Omega$ is in the form of upper and lower bounds for each design variable, then Problem \ref{prob:reconcilation} can be reduced to a simple projection $\bar \Xi_k$ to its closest bound. This is however not true owing to the coupling constraints~\eqref{eq:motor_design_constraints} and~\eqref{eq:con_magn_flux_yoke}. 

Finally, the motor design $\Xi^*$ is used to update the motor and manipulator models, based on which the trajectories $u^*_i$ for $1 \le i \le n$ can be determined.


\subsection{Differentiable Simulator} \label{subsec:differentiable_simulator}
Overcoming the computational challenge entails high efficiency of the simulator and optimization, which is governed by the number of time steps of integration and decision variables. We resort to CasADi \cite{Andersson.InPress2018} to implement a differentiable simulator of both the motor and robotic arm dynamics. The optimization benefits from this treatment because CasADi natively supports the derivation of analytical gradient formula as long as all functions are implemented in such a way that all outputs are differentiable w.r.t. inputs.

We wish to reduce the number of time steps, which compromises simulation accuracy and numerical stability. This conflict is worsened by the fact that the time scale of the motor dynamics is orders of magnitude faster than that of the arm dynamics. In virtue of the unsatisfactory performance for explicit numerical integration methods such as Runge-Kutta to deal with stiff dynamics, we adopt the direct collocation method to integrate the forward dynamics of both the motor and robotic arm in CasADi. This allows us to reduce the number of time steps. Below is a short sketch. Discretizing control over time interval $[0,t_f]$ yields
\begin{align}
    [u_{d}(t),\:u_{q}(t)] = [u_{d,w},\:u_{q,w}]_{12\times1},\\
    \:\:\:t\in[t_{w},t_{w+1}],\:\:\:w = 0,...,n_w-1,\nonumber
\end{align}
where the index $n_{w}$ represents the total number of discretization points. The vector $[u_{d,w},\:u_{q,w}]$ describes the voltage inputs for all six motors. The list of all collocation points is:
\begin{align}
    t_{w,j}=t_{w}+\Delta t_{w}\gamma_{j},\ \text{for}\:\:\:w=0,...,n_w-1;\:\:\:j=0,...,d.\nonumber
\end{align}
where $\Delta t_w=t_{w+1}-t_{w}$, $\gamma$ remarks the nodes, and the degree $d$ for the Lagrangian polynomials is set to $1$.

\subsection{ABA with Gearbox and Motor Inertia}
\label{sec:ABA_algorithm_with_gearbox_and_rotor_inertia}
We employ Alg.~\ref{alg:ABA_gearbox} for forward dynamics simulation, where $p$, $A$, $M$, $G$, $F_{tip}$, $\mathcal V$ and $Z$ are the biased force, screw axis, homogeneous transformation matrix, spatial inertia matrix, wrench at the tip, twist, and gear ratio, respectively. The ABA algorithm in \cite{Featherstone.2008} is extended to account for motor inertia and gear ratio. 
As in \cite{Lynch.2017}, we assume that the $(k+1)$th motor is mounted on the $k$th link, and thus the $k$th link inertia $G_{L_k}$ in Alg.~\ref{alg:ABA_gearbox} contains the stator of the $(k+1)$th motor. The motor rotor and gearbox in the $k$th axis is treated as an individual rigid body, with the inertia being denoted as $G_{R_k}$, and all related kinematic and dynamic quantities such as the twist are updated separately. The flexibility and backlashes of the gears are ignored. The spatial vector for gravitational acceleration is $g=-[0,0,0,0,0,9.81]^\top$. 

\begin{algorithm}[!h]
	\textbf{Inputs}: $\theta_{L_k},\dot{\theta}_{L_k} ,M_{L_k,L_{k-1}}, M_{R_k,L_{k-1}}, M_{R_k,L_k}, A_{L_k}, A_{R_k}, \tau_{L_k}, F_{tip}, G_{L_k}, G_{R_k}$ \;
	\textbf{Output}: $\ddot{\theta}_{L_k},\:\:k\in \{1,2,...,6\}$\;
	\textbf{Ensure}: $\mathcal V_{pre} = 0, \dot{\mathcal V}_0 = -g$ \;
	// update matrices and initialize inertia/biased forces of articulated bodies\;
	\For{$k=1$ \textbf{to} $6$}{
		Calculate homogeneous transformation matrix:\\
		$T_{L_k,L_{k-1}} = \mathrm{vTSE3}(-A_{L_k},\theta_{L_k})M_{L_k,L_{k-1}}$\;
		$T_{R_k,L_{k-1}} = \mathrm{vTSE3}(-A_{R_k},\theta_{L_k})M_{R_k,L_{k-1}}$\;
		Calculate adjoint matrices: $Ad_{T_{L_k,L_{k-1}}}, Ad_{T_{R_k,L_{k-1}}}$\;
        Calculate the twists:
        $\mathcal V_{L_k}=Ad_{T_{L_k,L_{k-1}}}\mathcal V_{pre} + A_{L_k}\dot{\theta}_{L_k}$\;
        $\mathcal V_{R_k}=Ad_{T_{R_k,L_{k-1}}}\mathcal V_{pre} + A_{R_k}\dot{\theta}_{L_k}$\;
		Calculate Lie bracket of twist:
		$ad_{\mathcal V_{L_k}}, ad_{\mathcal V_{R_k}}$\;
		Calculate:\\
		$\xi_{L_k} = ad_{\mathcal V_{L_k}}A_{L_k}\dot{\theta}_{L_k}, \xi_{R_k} = ad_{\mathcal V_{R_k}}A_{R_k}\dot{\theta}_{L_k}$\;
		Initialize the spatial inertia matrix:\\
		$\mathcal I^A_{L_k} = G_{L_k}, \mathcal I^A_{R_k} = G_{R_k}$\;
		Initialize the biased force:\\
		$p^A_{L_k} = ad_{\mathcal V_{L_k}} \mathcal I^A_{L_k} \mathcal V_{L_k}, p^A_{R_k} = ad_{\mathcal V_{R_k}} \mathcal I^A_{R_k} \mathcal V_{R_k}$\;
		Prepare for next iteration:
		$\mathcal V_{pre} = \mathcal V_{L_k}$\;
	}
	// update spatial inertia matrix and biased force\;
	\For{$k=6$  \textbf{to} $1$}{
		Calculate:\\
		$\mu_{L_k} = \tau_{L_k} - A_{L_k}p^A_{L_k}, \mu_{R_k} = \tau_{L_k} - A_{R_k}p^A_{R_k}$\;
        $D_k = 1/\left(A_{L_k}^\top \mathcal I^A_{L_k} A_{L_k} + A_{R_k}^\top \mathcal I^A_{R_k} A_{R_k}\right)$\;
		\If{$k>1$}{
			$\mathcal I^A_{L_{k-1}} = \mathcal I^A_{L_{k-1}}
			+
			Ad_{T_{L_k,L_{k-1}}}^\top (
			\mathcal I^A_{L_k}
			- \mathcal I^A_{L_k} A_{L_k} D_k A_{L_k}^\top \mathcal I^A_{L_k} ) Ad_{T_{L_k,L_{k-1}}}
			+
			Ad_{T_{R_k,L_{k-1}}}^\top (
			\mathcal I^A_{R_k}
			- \mathcal I^A_{R_k} A_{R_k} D_k A_{R_k}^\top \mathcal I^A_{R_k}) Ad_{T_{R_k,L_{k-1}}}$\;
			$p^A_{L_{k-1}} = p^A_{L_{k-1}}
			+ Ad_{T_{L_k,L_{k-1}}}^\top (
			p^A_{L_k}
			+ \mathcal I^A_{L_k}\xi_{L_k}
			+ \mathcal I^A_{L_k}A_{L_k}D_k(\mu_{L_k}
			- A_{L_k}^\top \mathcal I^A_{L_k}\xi_{L_k}))
			+
			Ad_{T_{R_k,L_{k-1}}}^\top (
			p^A_{R_k}
			+ \mathcal I^A_{R_k}\xi_{R_k}
			+ \mathcal I^A_{R_k}A_{R_k}D_k(\mu_{R_k} -
			 A_{R_k}^\top \mathcal I^A_{R_k}\xi_{R_k}))$
		} 
	} 
	// update joint accelerations\;
	$\dot{\mathcal V}_{k,pre} = \dot{\mathcal V}_0$
	\For{$k=1$  \textbf{to} $6$}{
		$\dot {\mathcal V}_i = Ad_{T_{L_k,L_{k-1}}} \dot{\mathcal V}_{k,pre}$\;
		$\ddot{\theta}_{L_k} = D_k(\tau_{L_k}
		- A_{L_k}^\top(
		\mathcal I^A_{L_k}(\dot{\mathcal V}_{L_k} + \xi_{L_k})
		+ p^A_{L_k})
		- A_{R_k}^\top(
		\mathcal I^A_{R_k}(\dot{\mathcal V}_{L_k} + \xi_{R_k})
		+ p^A_{R_k}))$\;
        $\dot{\mathcal V}_{L_k} = \dot{\mathcal V}_{L_k} + A_{L_k}\ddot{\theta}_{L_k} + \xi_{L_k}$\;
		Prepare for next iteration:
		$\dot{\mathcal V}_{k,pre} = \dot{\mathcal V}_{L_k}$\;
	} 
	\textbf{Return}: $\ddot{\theta}_{L_k}$, $\:\:k\in[1,2,...,6]$\;
	\caption{ABA for a 6DOF open-chain manipulator with gear boxes and motor inertia.}
	\label{alg:ABA_gearbox}
\end{algorithm}

\subsection{Discussions}

By directly transcripting Problem~\ref{prob:co-design} and solving it for all tasks at one time, we are able to obtain a solution converging to a local minimum. The proposed two-step process, albeit being much more computationally efficient, does not necessarily converge to a stationary point. As a result, the proposed algorithm yields sub-optimal design.

Let us take $n_w = 50$, which means that solving Problem~\ref{prob:co-design} for task $\mathcal T_i$ involves $m=1842$ decision variables. Solving Problem \ref{prob:co-design} for $n$ tasks at one time involves $n$ times more decision variables. So does the number of constraints. 
Since CasADi uses IPOPT to solve optimization problem based on gradient and Hessian, which possess a complexity of $O(m)$ and $O(m^2)$ with auto-differentiation, one can appreciate the computational advantage of the two-step process. 


This work can be readily extended to co-design of the geometric parameters of robot links. All robot link parameters appear in the model \eqref{eq:robot_dynamic_model} analytically, which implies that the cost function is an analytical function of the link parameters. Therefore, auto-differentiation remains viable.

\section{Simulation} \label{sec:simulation}
Simulation compares the results of applying three co-design methods to pick-and-place application with time-optimal and energy-optimal cost functions: 1) an empirical co-design for a task $\mathcal T_e$ with a medium stroke $\lambda=0.6$ and payload $M_p=2$ kg; 2) a worst-case co-design for a task $\mathcal T_w$ with the longest stroke $\lambda=1$ and maximum payload $M_p=4$ kg; and 3) the multi-task two-step based co-design using 25 tasks with varying strokes and payloads. Note that both the empirical and worst-case co-designs solve Problem \ref{prob:co-design} for tasks $\mathcal T_e$ and $\mathcal T_w$, respectively, by conducting motor design instead of merely motor selection as in literature, e.g.,~\cite{Pettersson.2009}. 

\subsection{Application-specific Task Set $\mathcal T$}
Without loss of generality, we specify the task set $\mathcal T$ by numerating initial states $\bm x_0$ over five choices
$$
\bm x_0 = \lambda\begin{bmatrix}\mathbf{1}_6 & \mathbf{0}_6\end{bmatrix}^\top, \text{ for } \lambda \in \{0.2,0.4,0.6,0.8,1\},
$$
and payload $M_p$ over five options $M_p \in \{0,1,2,3,4\} \text{ kg}.$
Here $\mathbf 1_6$ is a vector of ones in $\mathbb R^6$, and $\mathbf{0}_6$ is a vector of zeros in $\mathbb R^6$. For each choice of $\bm x_0$, we change $M_p$ (a solid ball) to be $0$, $1$, $2$, $3$, $4$ kg to define five tasks. Eventually, we obtain $25$ tasks to characterize the application, all of which  have the same final state $\bm x_f = \begin{bmatrix}\mathbf{0}_6 & \mathbf{0}_6\end{bmatrix}^\top$.
\subsection{Simulation Results}
\begin{figure}[thpb]
\centering
	\includegraphics[width=0.47\textwidth]{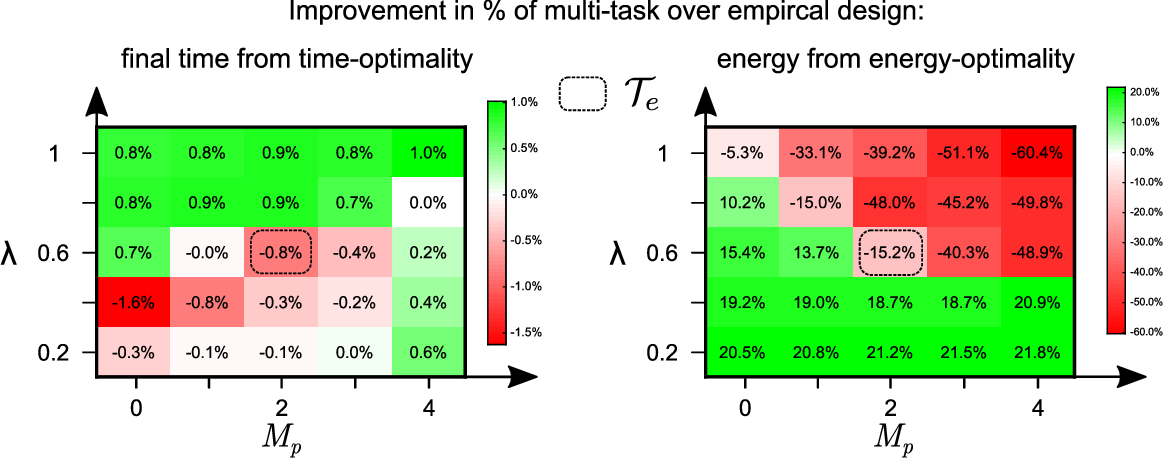}
	\caption{Percentages of improvement (green) of multi-task method over empirical design.}
\label{fig:07_heatmap_empircal_vs_codesign_for_alpha_0_time_opt}
\end{figure}

\begin{figure}[thpb]
\centering
	\includegraphics[width=0.47\textwidth]{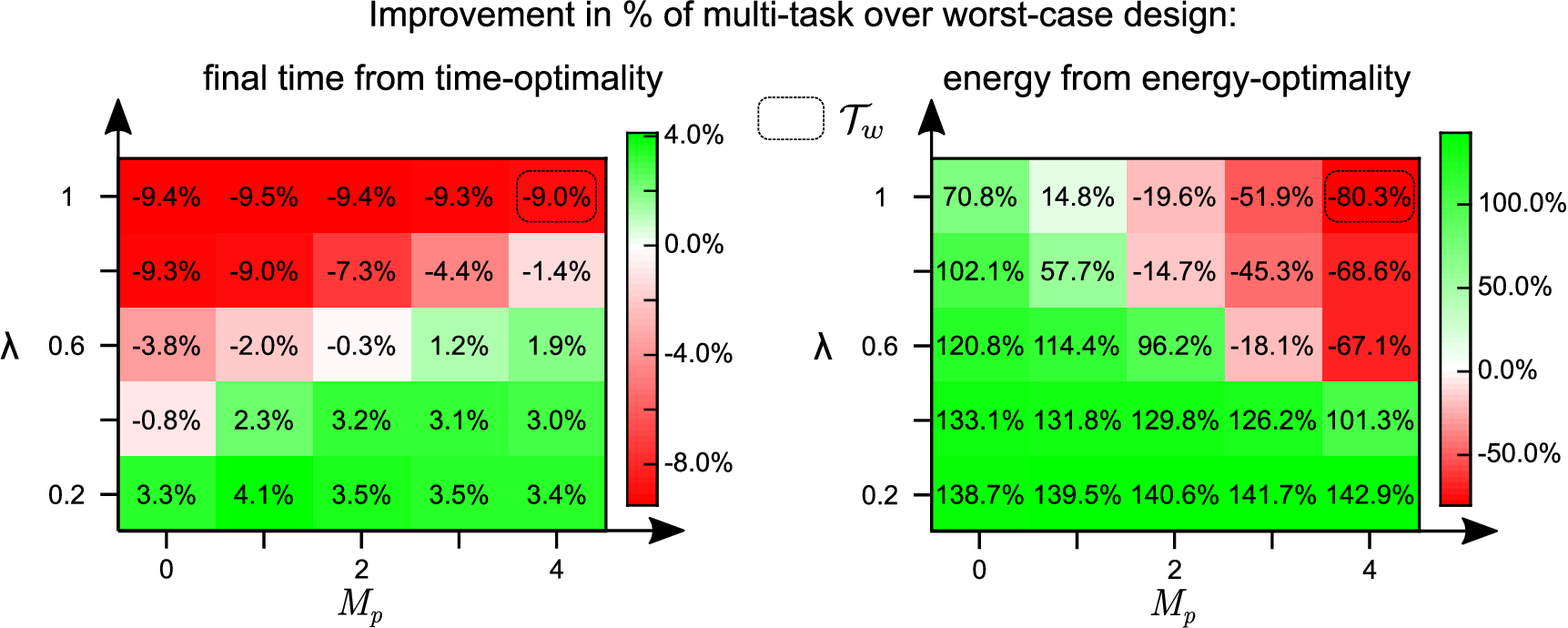}
	\caption{Percentages of improvement (green) of multi-task method over worst-case design.}
 \label{fig:08_heatmap_worstcase_vs_codesign}
\end{figure}

\begin{figure}[thpb]
\centering
	\includegraphics[width=0.48\textwidth]{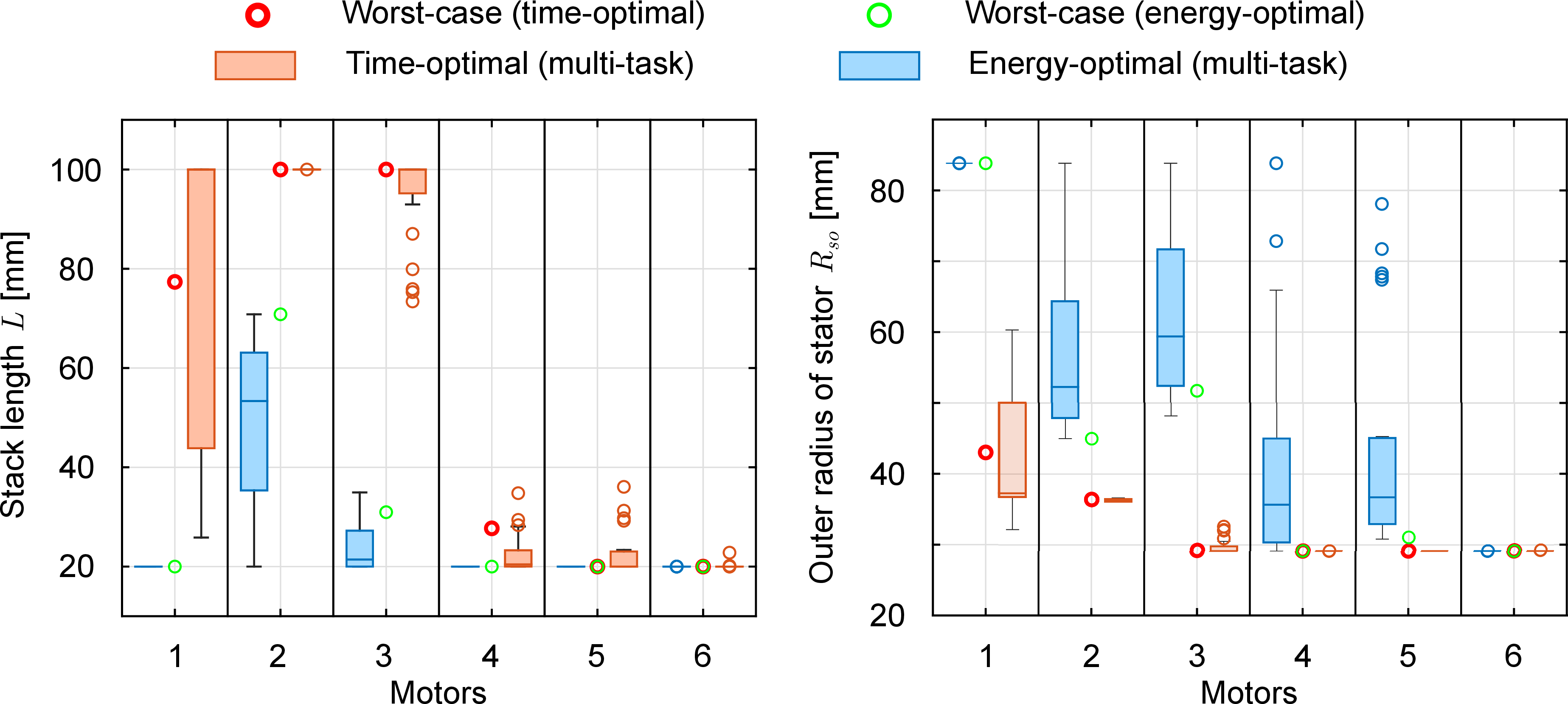}
	\caption{Stack length and outer stator radius of the motors from $25$ tasks (box whisker charts) versus worst-case design (circle): time-optimal (red) and energy-optimal (blue).}
 \label{fig:03_Benchmark_vs_Codesign_boxchart}
\end{figure}

Fig.~\ref{fig:07_heatmap_empircal_vs_codesign_for_alpha_0_time_opt} illustrates that the empirical outperforms the multi-task for neighboring tasks of $\mathcal T_e$ in terms of the final time and energy consumption, whereas the multi-task performs better once the discrepancy between test task and $\mathcal T_e$ gets relatively large (area in green). Note that task $\mathcal T_e$ is ideally picked for the range of $\lambda$ and $M_p$ when comparing it to the multi-task design where every task is equally weighted to derive the motor design. 
Fig.~\ref{fig:08_heatmap_worstcase_vs_codesign} shows the comparison between the multi-task and the worst-case design in terms of the final time (left plot) and energy consumption (right plot). As expected the worst-case design performs better for $\mathcal T_w$. However, considering the whole space of $\lambda$ and $M_p$, the multi-task method proves to provide good results, especially in terms of energy consumption for small initial conditions and light payloads. It should be noted that the final time- or energy consumption is seemingly more dependent on the initial conditions $\lambda$ than on the payload, because the payload puts less stress on motors than the manipulator does, which explains why the worst-case design performs better for long-stroke tasks.
The co-design results for 25 tasks are shown as box whisker charts in Fig.~\ref{fig:03_Benchmark_vs_Codesign_boxchart}, where red and blue mark the time and energy-optimal solutions. For illustration purposes, we provide the axial stack length $L$ and the outer stator radius $R_{so}$. 
The worst-case design results are shown as red and green circles for the time and energy-optimal designs, respectively. It can be seen that for the stack length of motor in axis $1$, $2$, and $3$, the multi-task yields time-optimal solutions close to the worst-case solution, with motor in axis $2$ exactly matching the worst-case design. A similar trend can be observed for the outer stator radius. Motor in axis $2$ revolves around the $y$-axis, counteracting the gravity. 
The time-optimal solution requires large accelerations. Therefore, the motor needs a long stack length to create high torques but a small outer stator radius to have a small inertia. Overall, the multi-task method offers more flexibility for future research to stochastically analyse different strokes and payloads to derive an optimal motor design. The empirical and worst-case design only considers one specific task where it could be seen that specifically in terms of energy-optimality the multi-task method outperforms these designs.
\section{Conclusions and Future Work}
\label{sec:conclusions}
This paper dealt with  joint optimization of motions and motors of a manipulator to achieve high performance for a particular application characterized by a set of tasks. We overcame the computational challenge of this multidisciplinary design problem by 1) proposing a two-step design process to resolve the scalability issue arising from multiple tasks of an application; 2) conducting analytical modeling of motor and manipulator to establish a differentiable mapping from motor design and motion to an objective function; 3) developing a differentiable simulator based on direct collocation to balance computation efficiency and simulation accuracy. Simulation was performed to validate the effectiveness of the application-oriented co-design paradigm. Future work includes 1) incorporating weight and friction of the gear box and perform co-design of the gear box, motor, arm, and control policy; and 2) exploring stochastic framework to obtain optimal motor design for multiple tasks.
\section*{Acknowledgment}
The authors thank Drs. Scott Bortoff, Diego Romeres, Marcel Menner, and Jinyun Zhang for helpful discussions.
\appendix
\renewcommand{\theequation}{A.\arabic{equation}}
\subsection{MEC Modeling for the SPMSM}
The MEC modeling is provided in this section. More details can be found in~\cite{Higuchi.2017}.
\subsubsection{Dimensional specifications}
\label{subsec:dimensional_specifications}
The cylindrical rotor weight can be calculated as $m_{rotor} = \rho_{iron}\pi R_{ro}^2 L$, and the inertia in its principal axes are $I_{xx} = \frac{1}{2}\rho_{iron}\pi R_{ro}^4 L$ and $I_{yy} = I_{zz} = \frac{1}{12} \rho_{iron}\pi R_{ro}^2 L \left(3R_{ro}^2 + L^2\right)$. The slot height is $h_{ss} = R_{so} - h_{sy} - R_{ro} - \delta - h_{tip}$. Assuming a rectangular tooth cross-sectional area (see Fig.~\ref{fig:0_Motor}), the slot width is $b_{ss} = \frac{A_{slot}}{h_{ss}}$ where
$$
    A_{slot} = \frac{\pi\left(\left(R_{so} - h_{sy}\right)^2 - \left(R_{ro} + \delta + h_{tip}\right)^2\right)}{Q} - w_{tooth} h_{ss}.
$$
\subsubsection{Stator and rotor weight}
\label{subsec:stator_rotor_weight}
For a given filling factor of $f_f=0.55$ the copper area can be calculated as $A_{cu} = A_{slot}f_f$. For a concentrated winding type and the assumption that one winding is a complete turn around a tooth, a single coil area is given by $A_{coil} = \frac{A_{cu}}{2n_s}$.
The stator and rotor weight can be calculated as follows:
\begin{align*}
    m_{stator} =& \rho_{iron} \pi R_{so}^2 L - \rho_{iron} \pi R_{ro}^2 L ...\nonumber\\
    &- \rho_{iron} A_{cu} L Q + \rho_{cu} A_{coil} L_{coil} n_s Q\\
    m_{rotor} =& \rho_{iron} \pi R_{ro}^2 L.
\end{align*}
\subsubsection{Resistance}\label{subsec:resistance}
The arc span $\tau_s$ per slot is given by $\tau_s = \frac{2\pi\left(R_{ro}+\delta\right)}{Q}$. The average length of the end-winding of the coil is $L_{end,av}=\frac{1}{2}\left(w_{tooth}\left(2-\frac{\pi}{2}\right) + \frac{\pi\tau_s}{2}\right)$ and the coil length $L_{coil}=2L + 2L_{end,av}$.
The resistance per tooth is given by $R_1 = \frac{n_s^2 \rho_e L_{coil}}{A_{slot}f_f}$, where $n_s$ is the number of windings per tooth. Finally, the phase resistance can be calculated as $R = \frac{q_1}{C^2}R_1$,
where $C$ is the number of coils connected in parallel and $q_1 = \frac{Q}{m}$ is the slots per phase (here $C=1$ and $m=3$) .
\subsubsection{Flux}
\label{subsec:flux}
Carter's coefficient $k_C$ is given as $k_C = \frac{t_{pitch}}{t_{pitch}-\gamma \delta}$ and the magnetic flux density across the gap is defined as $B_g$:
\begin{align*}
    \gamma = \frac{\left(\frac{b_0}{\delta}\right)^2}{5+\frac{b_0}{\delta}}, \quad
    t_{pitch} = \frac{2 \pi R_{ro}}{Q}, \quad B_g = B_r \frac{\frac{h_m}{\mu_r}}{\frac{h_m}{\mu_r} + \delta k_C}.
\end{align*}
The flux density of the first harmonics is $B_{g,1} = \frac{4}{\pi}B_g$, and the flux per tooth per single turn is $\Phi_1 = B_{g,1} \frac{2 \pi R_{ro}}{Q}L.$ Without skewness, we have the flux linkage $\Phi_m = k_w \frac{q_1}{C} n_s \Phi_1$, where $k_w = k_p k_d$ is the winding factor where $k_p = \sin\left(\frac{\pi P}{Q}\right)$ and $k_d = \frac{\sin\left(\frac{\pi}{6}\right)}{q_{pm}\sin\left(\pi/(6q_{pm})\right)}$.
$q_{pm} = \frac{q_1}{gcd(q_1,2P)}$ is te slots per pole per phase and $gcd(\cdot)$ means the great common divisor.
\subsubsection{Permeance}
\label{subsec:permeance}
The permeance of the magnetic path across the air gap and the slot opening, denoted by $p_g$ and $p_{so}$, respectively, can be calculated as follows:
\begin{equation*}
    p_g = \frac{\left(\frac{2 \pi R_{ro}\mu_0}{Q}\right) L} {\delta + \frac{h_m}{\mu_r}}, \quad p_{so} = \frac{\mu_0 h_{tip} L}{b_0}.
\end{equation*}
The permeance of the curved magnetic path from tip to tip is:
\begin{equation*}
    p_{tt} = \frac{\mu_0 \left(\delta + h_m\right) L}{\frac{\pi}{2}\left(\delta + h_m\right)}.
\end{equation*}
\subsubsection{Inductance}
\label{subsec:inductance}
The d- and q-axis inductance is given by
$L_{dq} = \frac{q_1}{C^2} n_s^2 L_1,$ where $L_1 = p_g + 3p_{so} + 3p_{tt}$ is the inductance per turn and per tooth.


\balance
\bibliographystyle{IEEEtran}


\end{document}